\documentclass[10pt, letter, onecolumn]{arxiv}

\usepackage{kantlipsum, lipsum}
\usepackage{dm-colors}
\usepackage{amsmath}
\usepackage{pstricks, pst-node}
\usepackage{verbatim}
\usepackage{multirow}
\usepackage{makecell}
\usepackage{scalerel}
\usepackage{booktabs}
\usepackage{enumitem}
\usepackage{xspace}
\usepackage{bm}
\usepackage{bbm}
\usepackage{mathtools}
\usepackage{soul}
\usepackage{epsfig}
\usepackage{graphicx}
\usepackage{amssymb}
\usepackage{colortbl}
\usepackage{arydshln}
\usepackage{mdframed}
\usepackage{csquotes}
\usepackage{setspace}
\usepackage{colortbl}
\usepackage{tabularx,ragged2e}
\usepackage{placeins}
\usepackage[dvipsnames]{xcolor}
\usepackage{bbm}
\usepackage{longtable}
\usepackage{lineno} % linenumber
\usepackage[hang,flushmargin]{footmisc}
\usepackage{nameref}
\usepackage{varioref}
\usepackage[pagebackref=false,breaklinks=false,%
            colorlinks=true,bookmarks=true,citecolor=ourdarkblue,%
            urlcolor=ourdarkblue,linkcolor=ourdarkblue]{hyperref}
\usepackage[noabbrev,capitalize]{cleveref}
\usepackage{etoc}
\usepackage{soul}

\usepackage{ifthen}

\graphicspath{{figures/}}

\newcommand{\model}[1]{{SAT}}
\newcommand{\samdataset}[1]{{SAT-DS}}

\newcommand{\eg}{{\textit{e.g.}}}

\newcommand{\ourMethodName}{EchoCare}
\newcommand{\datasetName}{EchoCareData}

\newcommand{\organNumber}{52}
\newcommand{\sliceNumber}{56}

\newcommand{\datasetNumber}{138}

\newcommand{\countryNumber}{20}
\newcommand{\continentNumber}{5}

\newcommand{\myparagraph}[1]{{\vspace{.5em} \noindent \bf #1}}

\definecolor{lightlightgray}{rgb}{200,200,200}

\title{\Large{A Fully Open and Generalizable Foundation Model for Ultrasound Clinical Applications}}

\author[1]{Hongyuan Zhang}
\author[1,2]{Yuheng Wu}
\author[3,4,$*$]{Mingyang Zhao}
\author[1,5]{Zhiwei Chen}
\author[6]{Rebecca Li}
\author[1,$*$]{Fei Zhu}
\author[1,2]{Haohan Zhao}
\author[7]{Xiaohua Yuan}
\author[8]{Meng Yang}
\author[9]{Chunli Qiu}
\author[9]{Xiang Cong}
\author[10]{Haiyan Chen}
\author[11]{\\ Lina Luan}
\author[12]{Randolph H.L. Wong}
\author[13]{Huai Liao}
\author[6]{Colin A Graham}
\author[7]{Shi Chang}
\author[9]{Guowei Tao}
\author[1]{\\ Dong Yi}
\author[1,4,14]{Zhen Lei}
\author[15]{Nassir Navab}
\author[16]{Sebastien Ourselin}
\author[1,17]{Jiebo Luo}
\author[1,14,16]{\\Hongbin Liu}
\author[1,4,14,$*$]{Gaofeng Meng}

\affil[1]{\justifying\normalsize Center for Artificial Intelligence and Robotics, Hong Kong Institute of Science \& Innovation, Chinese Academy of Sciences, Hong Kong, China \authorcr}
\affil[2]{\justifying\normalsize City University of Hong Kong, Hong Kong, China \authorcr}
\affil[3]{\justifying\normalsize State Key Laboratory of Mathematical Sciences, Academy of Mathematics and Systems Science, Chinese Academy of Sciences, Beijing, China \authorcr \vspace{0.1cm}}
\affil[4]{\justifying\normalsize University of Chinese Academy of Sciences, Beijing, China \authorcr \vspace{0.1cm}}
\affil[5]{\justifying\normalsize Division of Electronic Engineering, Faculty of Engineering, The Chinese University of Hong Kong, Hong Kong, China \authorcr}
\affil[6]{\justifying\normalsize Accident and Emergency Medicine Academic Unit, The Chinese University of Hong Kong, Hong Kong, China \authorcr}
\affil[7]{\justifying\normalsize Xiangya Hospital Central South University, Changsha, China \authorcr}
\affil[8]{\justifying\normalsize Hunan Frontline Medical Technology Co., Ltd, Changsha, China \authorcr}
\affil[9]{\justifying\normalsize Qilu Hospital of Shandong University, Jinan, China \authorcr}
\affil[10]{\justifying\normalsize Zhongshan Hospital of Fudan University, Shanghai, China \authorcr}
\affil[11]{\justifying\normalsize Shanghai Geriatric Medical Center, Shanghai, China \authorcr}
\affil[12]{\justifying\normalsize Division of Cardiothoracic Surgery, Department of Surgery, The Chinese University of Hong Kong, Hong Kong, China \authorcr}
\affil[13]{\justifying\normalsize Department of Pulmonary and Critical Care Medicine, The First Affiliated Hospital, Sun Yat-sen University, Guangzhou, China \authorcr}
\affil[14]{\justifying\normalsize State Key Laboratory of Multimodal Artificial Intelligence Systems, Institute of Automation, Chinese Academy of Sciences, Beijing, China \authorcr}
\affil[15]{\justifying\normalsize Computer Aided Medical Procedures, Technical University of Munich, Munich, Germany \authorcr}
\affil[16]{\justifying\normalsize School of Biomedical Engineering \& Imaging Sciences, King's College London, UK \authorcr}
\affil[17]{\justifying\normalsize Department of Computer Science, University of Rochester, USA \authorcr}
\affil[$*$]{\justifying\normalsize Corresponding authors}

\begin{document}
\begin{abstract}
\textbf{Abstract.} 
The inherent safety and versatility of ultrasound imaging have made it widely accessible in modern clinical settings for disease diagnosis and health management.
Artificial intelligence (AI) that can effectively learn ultrasound representations by integrating multi-source data holds significant promise for advancing clinical care.
However, the scarcity of large labeled datasets in real-world clinical environments and the limited generalizability of task-specific models have hindered the development of generalizable clinical AI models for ultrasound applications.
In this study, we present EchoCare, a novel ultrasound foundation model for generalist clinical use,
developed via self-supervised learning on our curated, publicly available, large-scale dataset EchoCareData.
EchoCareData comprises 4.5 million ultrasound images, sourced from over 23 countries across 5 continents and acquired via a diverse range of distinct imaging devices, thus encompassing global cohorts that are multi-center, multi-device, and multi-ethnic. Unlike prior studies that adopt off-the-shelf vision foundation model architectures, we introduce a hierarchical classifier into EchoCare to enable joint learning of pixel-level and representation-level features, capturing both global anatomical contexts and local ultrasound characteristics.
With minimal training, EchoCare outperforms state-of-the-art comparison models across 10 representative ultrasound benchmarks of varying diagnostic difficulties, spanning disease diagnosis, lesion segmentation, organ detection, landmark prediction, quantitative regression, imaging enhancement and report generation.
The code and pretrained model are publicly released, rendering EchoCare accessible for fine-tuning and local adaptation, supporting extensibility to additional applications.
EchoCare provides a fully open and generalizable foundation model to boost the development of AI technologies for diverse clinical ultrasound applications.
\end{abstract}
\maketitle

% \linenumbers
\tableofcontents

% ------------ SECTIONS ---------------------
\clearpage

\section{Introduction}

Ultrasound imaging stands as a cornerstone of modern medicine, celebrated for its unique combination of real-time assessment, cost-effectiveness, and inherent safety. This non-invasive and radiation-free modality allows for the dynamic visualization of physiological processes, securing its indispensable role in a wide range of clinical practices~\cite{wells2006ultrasound}. Despite these advantages, ultrasound diagnostic is heavily reliant on the skill of the sonographer and the specialized expertise required to interpret the complex, often subtle, visual information.
This inherent complexity, coupled with the ubiquity and versatility of ultrasound, has spurred significant interest in leveraging artificial intelligence (AI) to advance its use. As ultrasound imaging expands to new anatomical regions and clinical applications, there is a growing demand for versatile and generalizable AI models that can adapt to diverse clinical tasks and organs with minimal reliance on new labeled data. Meeting this demand will not only broaden the application of ultrasound analysis but also accelerate the deployment of smart healthcare solutions, making high-quality diagnostics more accessible and efficient.

Recent advances in foundation models (FM) using self-supervised learning have opened new frontiers in medical AI~\cite{zhou2023foundation,yan2025multimodal,chen2024towards,wang2024pathology,ma2025fully,ma2025generalizable}. These models learn general-purpose feature representations directly from raw data, eliminating dependence on extensive expert annotations. Upon completion of pretraining, these models can be effectively adapted to a wide array of downstream clinical tasks with minimal or no additional fine-tuning.
This paradigm represents a significant advantage over conventional medical AI approaches,
which are typically limited to specific anatomical structures or require extensive retraining when adapted to each new clinical application.
However, pretraining of foundation models requires large-scale and diverse datasets, making data acquisition and rigorous curation essential for developing clinically reliable and generalizable systems.

Building on the success of vision foundation models, researchers have started adapting these approaches to ultrasound imaging analysis~\cite{jiao2024usfm,kang2025urfm,xu2024whole}. Although initial results show promise, several critical challenges could limit their potential clinical impact. First, the scale of available ultrasound datasets remains relatively small, undermining the reliability of models for clinical deployment. Moreover, much of the pretraining data employed in previous studies is private, creating barriers to reproducibility, broader research and application. 
Second, current collections often focus on narrow anatomical regions, which is insufficient to fully capture the diversity of whole-body regions. This limitation restricts their utility in comprehensive clinical workflows.
Third, most approaches rely on off-the-shelf vision foundation model frameworks~\cite{he2022masked}, failing to systematically explore network architecture optimizations tailored to the morphological complexity and spatial hierarchies of anatomical structures.
This oversight limits the model's ability to capture anatomical relationships across scales and organs during pre-training.
Finally, existing research mainly focuses on a few downstream tasks such as image classification or segmentation, leaving open questions about model capabilities for more diverse clinical applications.

In this work, we introduce EchoCare, a novel foundation model for ultrasound images, accompanied by a systematic investigation of its utility across a diverse spectrum of clinical tasks. EchoCare is pre-trained on EchoCareData, our newly curated large-scale and openly accessible dataset comprising 4.5 million ultrasound images. Collected from multi-center, multi-device, multi-modality, and multi-ethnic global sources, \datasetName~ensures diverse data representation. \datasetName~covers 9 major regions and \organNumber~anatomical organs of the human body, supporting models pretrained on it to generalize effectively across comprehensive whole-body ultrasound clinical applications.
We have also optimized the architecture of the vision foundation model to better capture hierarchical anatomical structures, from broad ultrasound regions (\eg, abdomen) to specific organs (\eg, liver, kidney), enabling the model to mimic human-like clinical diagnostic reasoning.
Extensive evaluations across eight categories of core ultrasound clinical tasks of varying diagnostic difficulties, such as lesion segmentation, organ detection, disease diagnosis, and quantitative regression, reveal that \ourMethodName~significantly outperforms state-of-the-art general-domain foundation models, underscoring the critical need for ultrasound-specific models.
Compared with leading ultrasound-focused foundation models,
\ourMethodName~also demonstrated superior performance, highlighting the advantages of pretraining on large, diverse data. 
We will release both \ourMethodName~and the \datasetName~to promote clinical AI development in ultrasound images upon publication. 

\begin{figure}[!t]
    \centering
    \includegraphics[width=0.98\linewidth]{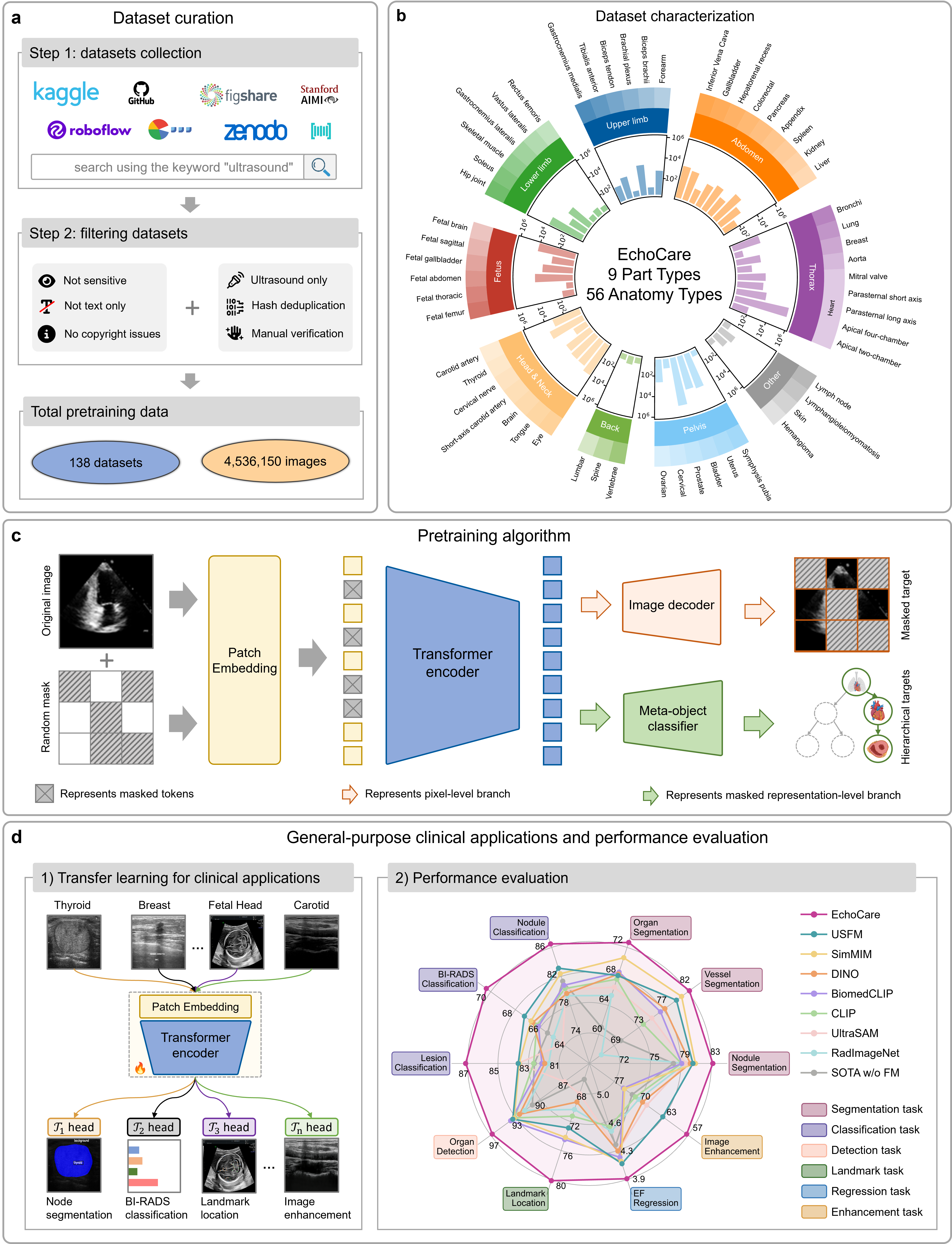}
    \centering
    \caption{\textbf{Overview of this study.} Caption on next page.}
    \label{fig:fig_overview}
\end{figure}
\begin{figure}
\caption*{(Previous page.) \textbf{Figure 1: Overview of this study.}
\textbf{a.} The ultrasound data from over \countryNumber~countries and \continentNumber~continents are collected, encompassing over 4 million ultrasound images.
\textbf{b.} The constructed ontology shows a hierarchy of object types that are used to unify semantic concepts across datasets.
Bar plots showing the number of images containing that object type.
\textbf{c.} Flowchart of EchoCare. \ourMethodName~takes a masked image as input and then outputs the reconstructed ultrasound image.
To capture both global anatomical contexts and local ultrasound characteristics, \ourMethodName~also incorporates a novel hierarchical classifier branch.
\textbf{d.} Performance evaluation across diverse clinical applications, spanning different organs and a wide range of diagnostic difficulties. EchoCare achieved state-of-the-art performance across all downstream clinical tasks.
Image was created with BioRender.com.}
\end{figure}

\section{Results}\label{sec:results}

\begin{figure}[!t]
    \centering
    \includegraphics[width=1.0\linewidth]{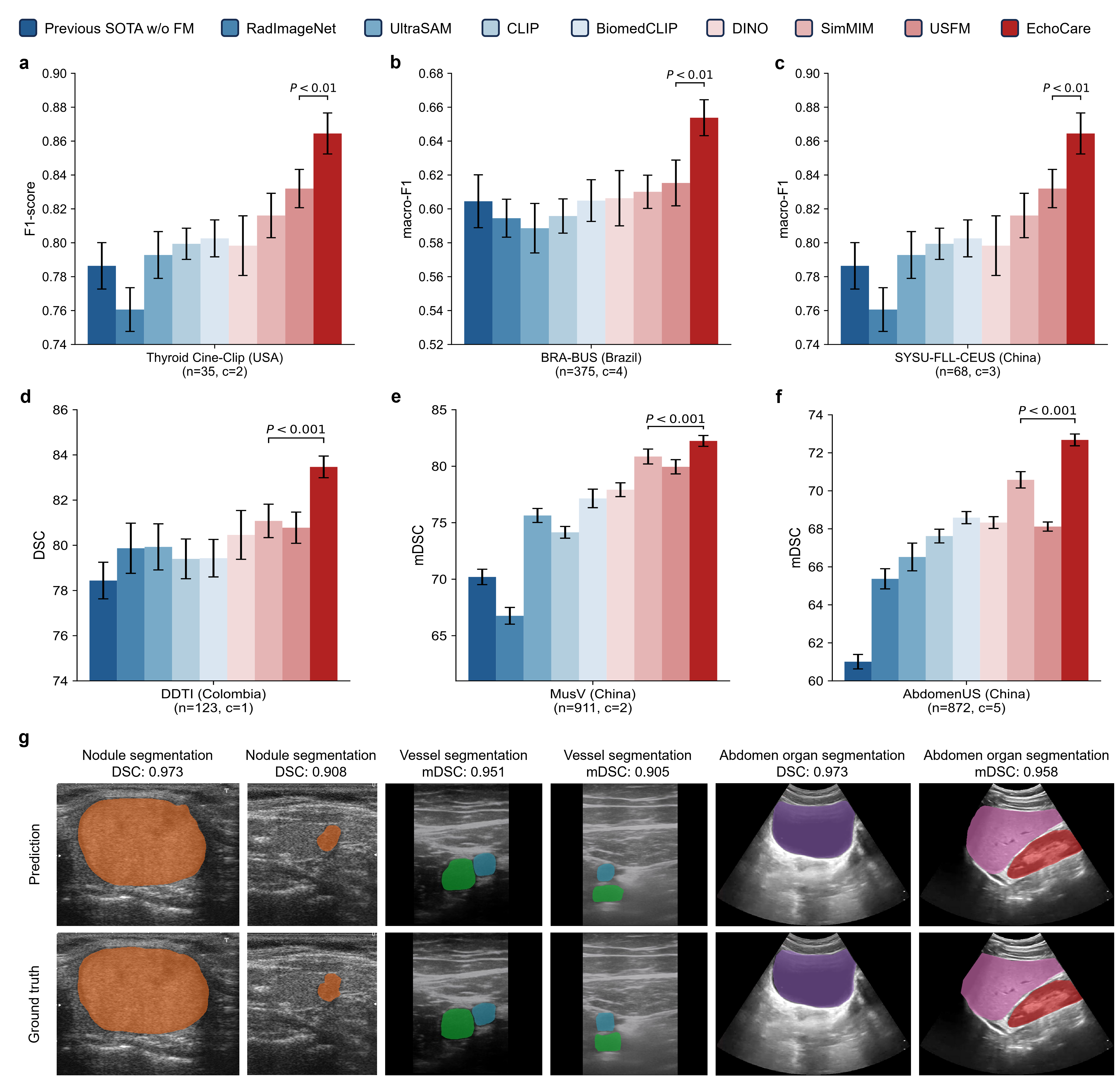}
    \caption{
     \textbf{Evaluation on disease diagnostic classification and anatomical segmentation.}
      \textbf{a-f.} \ourMethodName~consistently outperforms previous state-of-the-art (SOTA) models (w/o FM) and other existing foundation models (RadImageNet~\cite{mei2022radimagenet}, UltraSAM~\cite{meyer2024ultrasam}, CLIP~\cite{radford2021learning}, BiomedCLIP~\cite{zhang2025multimodal}, DINO~\cite{zhangdino}, SimMIM~\cite{xie2022simmim}, USFM~\cite{jiao2024usfm}) across different classification and segmentation tasks. Specifically, for classification, we evaluate on benign-malignant classification of thyroid nodules (\textbf{a}), breast tumor BI-RADS grading (\textbf{b}) and diagnosis of focal liver lesions in abdominal ultrasound (\textbf{c}). For segmentation, we evaluate on thyroid node segmentation (\textbf{d}), arterial-venous vessel segmentation (\textbf{e}), and the abdomen multi-organ segmentation (\textbf{f}).
     The two-sided Wilcoxon signed-rank test was used to assess the statistical differences between \ourMethodName~and the second-best model.
     \textbf{g.} Six examples comparing the segmentation results by \ourMethodName~and the ground truth.
     }
    \label{fig:fig_result_part1}
\end{figure}

\subsection{The largest ultrasound dataset \datasetName}
We establish so far the largest public ultrasound image dataset EchoCareData (Fig. \ref{fig:fig_overview}a,b), integrating \datasetNumber~ultrasound image datasets from over \countryNumber~countries and \continentNumber~continents. Encompassing multiple body organs, scanning devices, imaging modalities, and racial backgrounds (Fig. \ref{fig:fig_overview}b), the dataset is designed to ensure data diversity and enhance the generalization of pretrained models across diverse clinical applications.
\datasetName~adheres to rigorous cohort inclusion and exclusion protocols to ensure high quality, including manual removal of sensitive and non-ultrasound images, as well as text cleaning (Fig. \ref{fig:fig_overview}a). Using a clinical anatomy system, we generated canonical categorical labels for each image. The dataset’s ontology comprises eight representative clinical regions including head, chest, abdomen, limbs, back, fetus, dorsum, pelvis, and an ``other'' category, with a hierarchical structure spanning \organNumber~meta-object types (\eg, cardiac ventricle) to \sliceNumber~specific anatomic types (\eg, left cardiac ventricle), mirroring clinical diagnostic workflows. Moreover, an additional manual inspection was performed by randomly sampling 100 images from each class in \datasetName~to validate correctness. In total, \datasetName~comprises over 4.5 million distinct image-class tuples, spanning five imaging modalities (B-mode, CEUS, Dropper, M-mode, and Elastography), establishing it as a large-scale, diverse resource for clinical ultrasound care.

\subsection{Architecture and pre-training protocol of the foundation model EchoCare}
Building on EchoCareData, we pretrained \ourMethodName~(Fig. \ref{fig:fig_overview}c), a novel vision foundation model for ultrasound imaging, and applied it to a suite of clinical tasks. \ourMethodName~employs a modular design based on an extended self-supervised Masked AutoEncoder (MAE) architecture for representation learning, comprising an image encoder to encode input ultrasound image features and two decoders: an image decoder to reconstruct images from sparse patches and an anatomy-classifier decoder for joint learning of hierarchical anatomic features (Fig. \ref{fig:fig_overview}c). Unlike prior medical foundation models that directly adopt off-the-shelf MAE structures or focus solely on local pixel-level prediction, we introduce a novel representation-level prediction branch, the anatomy-classifier, into the MAE framework. This branch learns global and hierarchical anatomical relationships from body regions to organs to anatomic structures, mirroring clinical diagnostic workflows. For example, the anatomy-classifier predicts pathways such as ``Thorax$\rightarrow$Heart$\rightarrow$Apical two-chamber'' and ``Thorax$\rightarrow$Heart$\rightarrow$Apical four-chamber''. Leveraging the inherent hierarchical organization of the anatomy system, this high-level classification process evolves naturally without human intervention. By integrating local pixel-level and global representation-level features, \ourMethodName~enhances the encoder’s ability to interpret ultrasound images, thereby boosting downstream clinical applications.
In the following sections, we demonstrate its versatility and generalization to diverse ultrasound clinical tasks.

\subsection{EchoCare exhibits excellent performance across diverse ultrasound applications}
We systematically evaluated EchoCare diagnostic performance on 11 clinical applications across 8 task types (Fig. \ref{fig:fig_result_part1}, Fig. \ref{fig:fig_result_part2} and Fig. \ref{fig:fig_report}).
These datasets cover tasks ranging from binary diagnosis task to multi-class classification,
single-class tumor segmentation to abdominal multi-organ segmentation,
as well as ultrasound image enhancement, fetal landmark localization, organ detection, cardiac ejection fraction regression, and clinical report generation.
We compare EchoCare with pervious state-of-the-art (SOTA) task-specific models (w/o FM) and seven representative foundation models: RadImageNet~\cite{mei2022radimagenet}, UltraSAM~\cite{meyer2024ultrasam}, CLIP~\cite{radford2021learning}, BiomedCLIP~\cite{zhang2025multimodal}, DINO~\cite{zhangdino}, SimMIM~\cite{xie2022simmim}, USFM~\cite{jiao2024usfm}. 
Each model is fully fine-tuned on the task-specific dataset and evaluated with their corresponding metrics.
EchoCare consistently outperformed all other models, achieving significant improvements on 10 clinical tasks. 
These results validate the effectiveness of Echocare.
The domain-specific analyses of the experimental results are as follows.

\subsubsection{Disease diagnostic classification}
Disease diagnostic classification represents a pivotal clinical application of vision foundation models. High-performance ultrasound foundation models can substantially enhance the accuracy of disease lesion classification, mitigate false-positive decisions, and thereby reduce patient anxiety and costs.
To demonstrate the utility in clinical decision-making,
EchoCare was validated across three distinct diagnostic classification applications:
1) benign-malignant classification of thyroid nodules; 2) breast tumor BI-RADS grading; and 3) diagnosis of focal liver lesions in abdominal ultrasound. 

EchoCare achieved leading performance across all the evaluated classification tasks (Fig. \ref{fig:fig_result_part1}a-c). Specifically,
\ourMethodName~achieved an AUC (Area Under the ROC Curve) of 86.48\%  and an F1-score of 87.45\%  on the thyroid nodule dataset (Fig. \ref{fig:fig_result_part1}a); 70.36\%  accuracy and 65.38\%  macro-F1 on breast BI-RADS grading (Fig. \ref{fig:fig_result_part1}b); and 87.12\%  accuracy and  83.44\%  macro-F1 for focal liver lesions (Fig. \ref{fig:fig_result_part1}c). Compared with the second-best model (USFM~\cite{jiao2024usfm}), \ourMethodName~outperformed by average margins of \(3.35\%\) (AUC) and \(4.25\%\) (F1-score) on thyroid nodules, \(3.09\%\) (accuracy) and \(3.85\%\) (macro-F1) on breast BI-RADS, and \(3.45\%\) (accuracy) and \(3.98\%\) (macro-F1) on focal liver lesions. These findings highlight \ourMethodName~as a powerful foundation model capable of learning discriminative image representations, demonstrating great potential in distinguishing subtle differences between hepatocellular carcinoma, hemangiomas, and focal nodular hyperplasia. These lesions often exhibit overlapping sonographic appearances, which is a key challenge in achieving accurate manual diagnosis. Collectively, the experimental results confirm that EchoCare serves as a reliable diagnostic auxiliary tool, advancing ultrasound-based disease diagnostic classification and accelerating the clinical decision-making process.

\subsubsection{Anatomical segmentation}  
Accurate segmentation in ultrasound images enables clinicians to characterize morphological features (\eg, size, shape) and detect pathological abnormalities (\eg, neoplastic lesions), which is fundamental for treatment planning and prognosis assessment.
We evaluated different foundation models on three representative ultrasound clinical benchmarks for anatomical segmentation:
the DDTI dataset~\cite{pedraza2015open} for thyroid node segmentation, the Mus-V dataset~\cite{geng2024force} for arterial-venous vessel segmentation, and the abdomen multi-organ segmentation.

Compared with existing methods,
EchoCare achieved significantly higher performance (Fig. \ref{fig:fig_result_part1}d-f), surpassing the next best Dice Similarity Coefficient (DSC) by 2.09\% and Normalized Surface Dice (NSD) by 2.26\% in the thyroid nodule segmentation task, mDSC by 1.36\% and mNSD by 1.03\% in the vessel segmentation task (Fig. \ref{fig:fig_result_part1}d),
mDSC by 2.10\% and mNSD by 4.36\% in the multi-organ segmentation task (Fig. \ref{fig:fig_result_part1}f). In the vascular segmentation task, EchoCare outperforms the second-ranked model (SimMIM) by a remarkable margin (Fig. \ref{fig:fig_result_part1}e), which holds clinical significance for real-time vascular interventions (\eg, coronary procedures).
Furthermore, EchoCare also surpassed previous SOTA without FM architecture (SwinUNETR~\cite{hatamizadeh2021swin}) in benchmark evaluations.
We showed examples comparing EchoCare segmentation and the ground truth across multiple organs, demonstrating the generalizability of EchoCare (Fig. \ref{fig:fig_result_part1}g).
The strong performance of EchoCare on segmentation tasks represents a breakthrough for comprehensive abdominal assessments, as clinicians require simultaneous visualization of the liver, pancreas, and kidneys to detect pathological relationships (\eg, liver lesions compressing adjacent organs). Such consistency across single-organ, vascular, and multi-organ tasks underscores,
EchoCare’s capacity to learn generalizable ultrasound features for effective task adaptation.

\begin{figure}[!t]
    \centering
    \includegraphics[width=\linewidth]{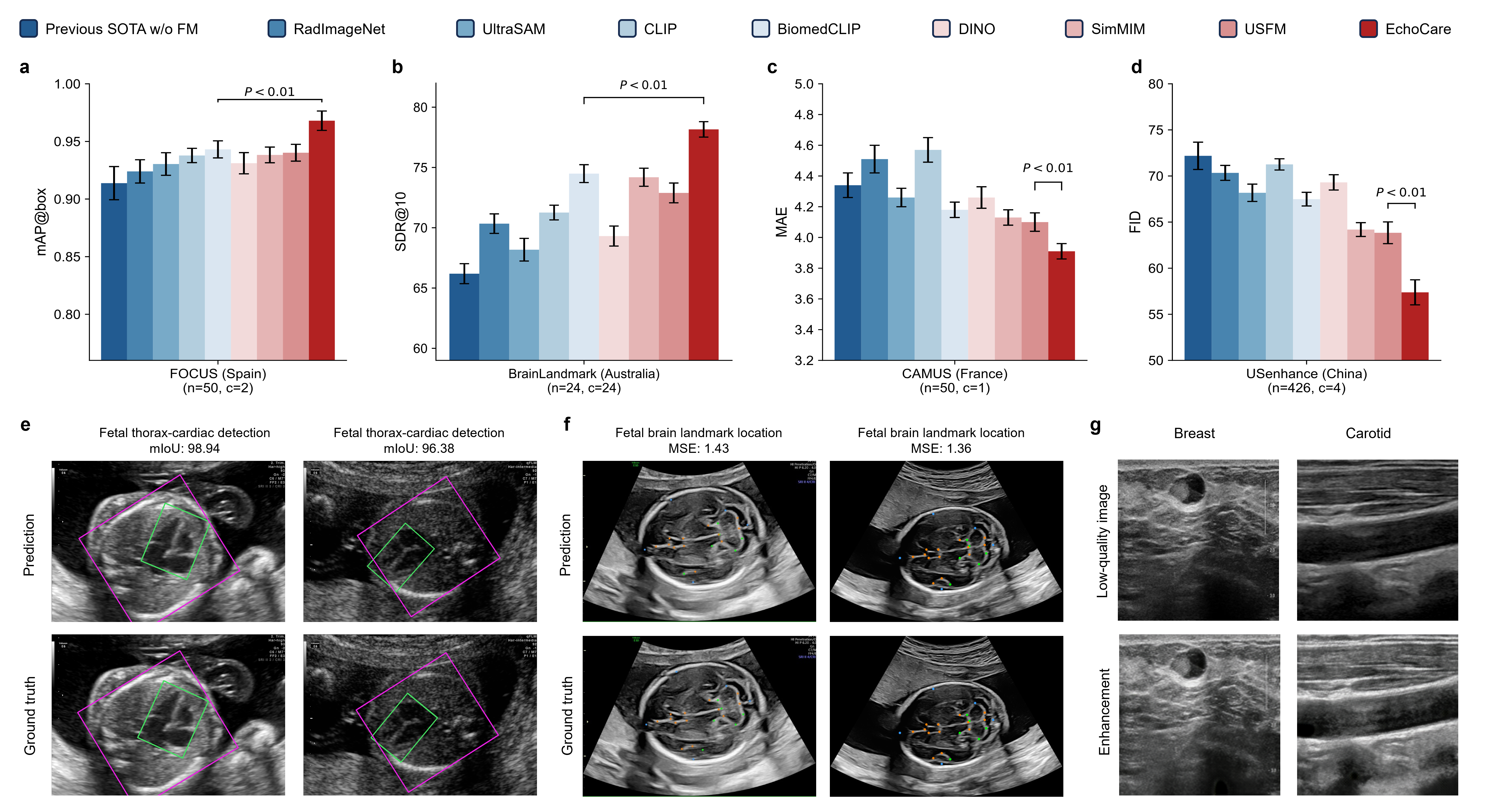}
    \caption{
     \textbf{Evaluation on organ detection, landmark prediction, fraction regression and imaging enhancement.}
     \textbf{a-d.} \ourMethodName~consistently outperforms pervious SOTA task-specific models (w/o FM) and existing foundation models across different tasks: organ detection (\textbf{a}), landmark prediction (\textbf{b}), fraction regression (\textbf{c}) and imaging enhancement (\textbf{d}). The two-sided Wilcoxon signed-rank test was used to assess the statistical differences between \ourMethodName~and the second-best model.
     \textbf{e-f.} Six examples comparing the detection, location and imaging enhancement results by \ourMethodName~and the ground truth.
     }
    \label{fig:fig_result_part2}
\end{figure}

\subsubsection{Fetal cardiac organ detection}
Fetal congenital heart disease (CHD) is a leading cause of infant mortality from birth defects, with an incidence reaching up to 8-10 cases per 1,000 live births~\cite{bravo2018prenatal,Wu2025}. Survival rates, requirement for intensive medical care, and risk of developmental disabilities are contingent on the accuracy and timeliness of diagnosis. Thus, early and precise prenatal sonographic diagnosis of CHD has been shown to reduce the risk of perinatal morbidity and mortality. The four-chamber view in fetal echocardiography is a unique and essential tool for assessing CHD. Diagnosis in this view relies on the cardiothoracic diameter ratio (CTR), a biometric defined as the ratio of thoracic to cardiac short-axis diameters. Therefore, detecting thoracic and cardiac regions from four-chamber echocardiograms is a critical step for CTR analysis and represents a foundational step in CHD diagnosis.

In this study, we evaluated the performance of \ourMethodName~against seven state-of-the-art foundation models and one previous SOTA model (Rotated Faster R-CNN~\cite{ren2016faster}) for fetal thorax and cardiac organ detection using the publicly available FOCUS dataset (300 four-chamber fetal echocardiography ultrasound images).
\ourMethodName~outperformed all other models significantly (Fig. \ref{fig:fig_result_part2}a).
Specifically, it $97.26\%$ AP (Average Precision) for thoracic object detection, and $96.11\%$ AP for cardiac object detection. Compared to the top ImageNet-based model (Rotated Faster-RCNN), \ourMethodName~showed even larger margins ($5.42\%$ higher mAP). This outcome underscores that ultrasound, specific pretraining, distinct from natural image pretraining, more effectively captures the domain-specific knowledge inherent to ultrasound imaging. Benefiting from its high detection accuracy, \ourMethodName~also ranked first in CTR measurement accuracy ($94.42\%$). We provide examples to visualize the detection results of EchoCare and demonstrate the superior performance by comparing with the ground truth (Fig. \ref{fig:fig_result_part2}e). These comprehensive results demonstrate that \ourMethodName~has the potential to enhance and accelerate prognosis prediction for CHD in ultrasound clinical practice.

\subsubsection{Fetal brain landmark predication}

Brain development involves progressive structural changes from early embryonic stages to several months after birth. Identifying fetal brain structures in ultrasound images enables assessment of cortical and subcortical gray matter changes,  serving as a valuable tool for detecting developmental abnormalities. However, manual landmark identification is labor-intensive, time-consuming, and prone to intra- and inter-rater inconsistency.

To address this challenge, we evaluated the performance of \ourMethodName~against other models for predicting fetal brain landmarks using the publicly available BrainBenchmark dataset~\cite{cabezas2024benchmark} (104 2D fetal brain ultrasound images acquired at 20-20.6 weeks of gestation from 70 pregnant women). \ourMethodName~outperformed all foundation models significantly (Fig. \ref{fig:fig_result_part2}b), achieving a notably lower average MSE (7.71) compared to the second-best model. \ourMethodName~also dominated in successful detection rate (SDR) across all pixel thresholds: at $\tau=2.0$ pixels, it achieved a SDR of $36.27\%$, (surpassing the second-best model SimMIM’s $30.24$); at $\tau=4.0$ pixels, it achieved $49.13\%$ (substantially exceeding SimMIM’s $42.87\%$); and at $\tau=10.0$ pixels, it achieved $80.16\%$ (versus BiomedCLIP’s $74.49\%$). We also provide examples to visualize the landmark prediction results of EchoCare and compare with the ground truth (Fig. \ref{fig:fig_result_part2}f). These results highlight EchoCare’s superiority in ultrasound-based landmark prediction, positioning it as a promising tool for automated fetal brain assessment.

\subsubsection{Cardiac ejection fraction regression}
The assessment of Left Ventricular Ejection Fraction (LVEF) is one of the most important manners in the evaluation of cardiac function. It quantifies the proportion of blood ejected from the left ventricle relative to its total end-diastolic volume. In clinical settings, accurate measurement of LVEF is pivotal to the early diagnosis of both congenital and acquired cardiovascular disorders, informs therapeutic decision-making, and enables robust prognostic stratification.

After observing the superior performance of \ourMethodName~across a range of ultrasound clinical tasks, we further evaluated it on the LVEF regression task using the CAMUS benchmark dataset~\cite{leclerc2019deep}. This dataset encompasses 2D apical four-chamber and two-chamber view sequences from 500 patients. Model performance is quantified by mean absolute error (MAE) with standard error. \ourMethodName~exhibited superior performance and outperformed the other 8 competing approaches (Fig. \ref{fig:fig_result_part2}c). It achieved the lowest MAE of $3.91$, surpassing the second-best pretrained model (USFM) by a $19\%$ reduction in MAE. Notably, it significantly outperformed the echo-specific state-of-the-art model (EchoMEM), with a significant $43\%$ reduction in MAE. These contributions underscore the potential of \ourMethodName~to advance cardiac LVEF regression and its applicability in real-world clinical workflows.

\subsubsection{Low-quality imaging enhancement}
High-quality ultrasound imaging is critical for the accurate identification of anatomical structures and disease diagnosis. However, ultrasound examinations using handheld or low-end devices often yield suboptimal images that compromise clinical diagnosis, particularly in resource-limited hospitals or regions. Enhancing such low-quality ultrasound images using AI technologies, for example, through improved contrast, sharpness, and signal-to-noise ratio, alongside noise reduction, could provide a cost-effective alternative to high-end scanners. This approach may also promote the wider adoption of portable ultrasound systems, offering substantial clinical benefits and ultimately improving patient outcomes. 

We evaluated \ourMethodName~on the low-quality ultrasound image enhancement task using the USenhance benchmark dataset~\cite{Guo2023}, which encompasses real-world clinical scans from 109 patients across five anatomical regions: thyroid, kidney, liver, breast, and carotid artery. \ourMethodName~was compared with 8 models, including previous SOTA model (EnlightenGAN~\cite{jiang2021enlightengan}), ultrasound-based models (RadImageNet~\cite{mei2022radimagenet}, UltraSAM~\cite{meyer2024ultrasam}), image-text multimodal models (CLIP~\cite{radford2021learning}, BiomedCLIP~\cite{zhang2025multimodal}), and self-supervised frameworks (DINO~\cite{zhangdino}, SimMIM~\cite{xie2022simmim}, USFM~\cite{jiao2024usfm}). Consistent with previous findings, \ourMethodName~outperformed all competing models across four metrics: NIQE, BRISQUE, PIQE, and FID (Fig. \ref{fig:fig_result_part2}d). Specifically, \ourMethodName~achieved mean NIQE, BRISQUE, PIQE, and FID values of $6.35\%$, $17.62\%$,  $30.16\%$, and $57.38\%$, respectively.
These visualizations (Fig. \ref{fig:fig_result_part2}g) further demonstrate the superior image quality enhancement ability of \ourMethodName.
These results demonstrate that \ourMethodName~can effectively enhance low-quality ultrasound images, highlighting the potential of AI for practical clinical applications in resource-limited settings.

\begin{figure}[!t]
    \centering
    \includegraphics[width=0.98\linewidth]{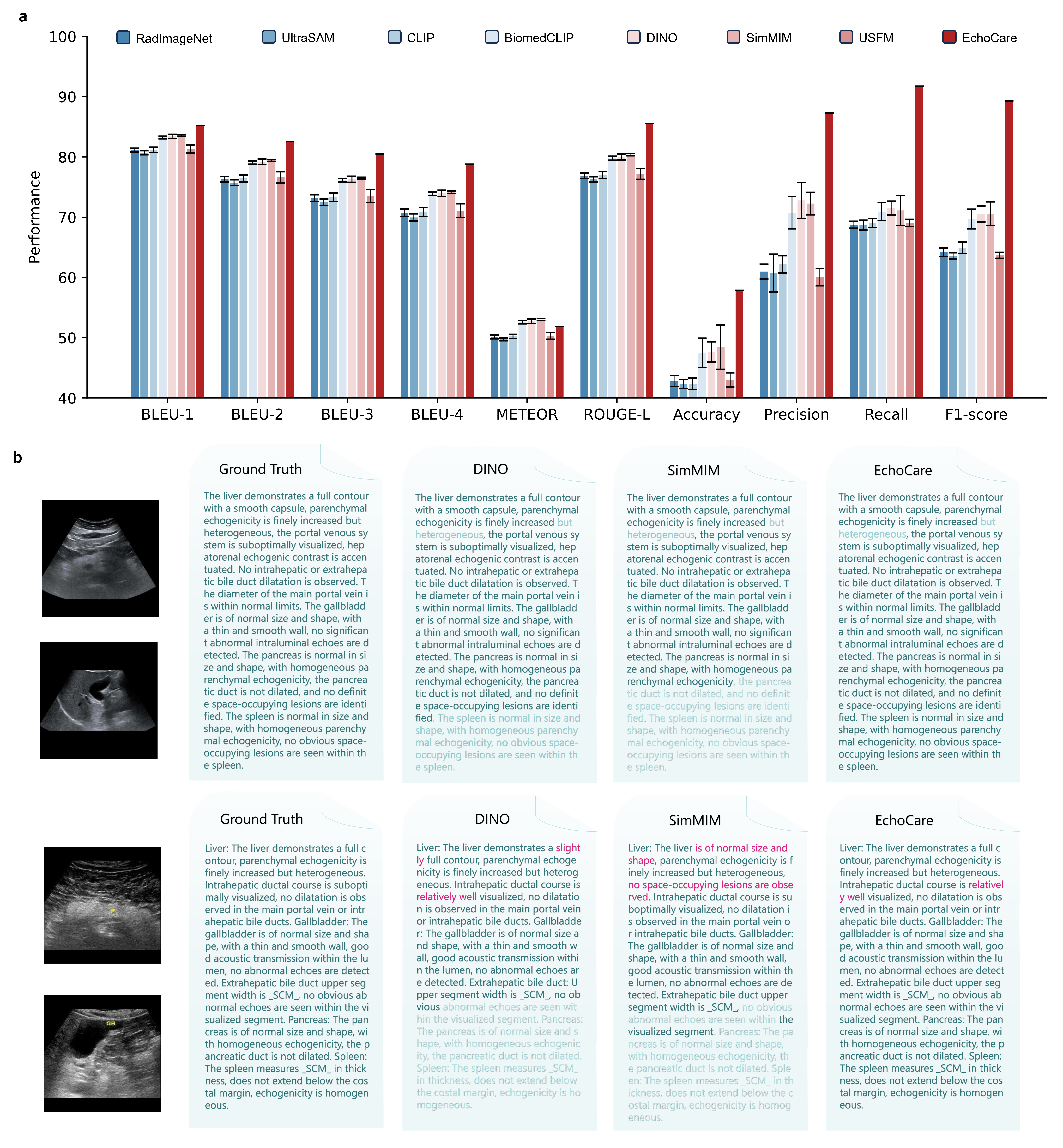}
    \caption{
     \textbf{Clinical report generation on USData~\cite{li2024ultrasound} Liver dataset.}
     \textbf{a.} Performance (\%) comparison of models on the USData~\cite{li2024ultrasound} Liver dataset using six language metrics (BLEU-1 to BLEU-4, METEOR, and ROUGE-L) and four classification metrics (Accuracy, Precision, Recall, and F1-score). Error bars denote standard deviation across multiple runs.
     \textbf{b.} Example reports generated by the two strongest baseline models (DINO~\cite{zhangdino} and SimMIM~\cite{xie2022simmim}) and \ourMethodName, compared against ground truth reports. Deep blue text indicates exact matches, light-colored text denotes missing segments, and vivid purple highlights over-generated content.
     }
    \label{fig:fig_report}
\end{figure}
\subsubsection{Clinical report generation}
Report generation is essential for healthcare system, providing critical information to clinicians and patients for the diagnosis, prognosis, and treatment planning of a wide range of medical applications. Traditionally, ultrasound reports are written manually by sonographers, which is time-consuming and prone to inter-observer variability.
Recent advancements in natural language processing and medical image analysis have enabled the development of automated ultrasound report generation systems.

To evaluate the effectiveness of our developed foundation model in ultrasound report generation, we integrate \ourMethodName~into an existing Transformer-based encoder–decoder report generator, where the input is the global visual features extracted from ultrasound images. The integrated model is then fine-tuned on the USData Liver dataset~\cite{li2024ultrasound}, which contains paired ultrasound images and corresponding expert-written reports. 
The experimental results (Fig. \ref{fig:fig_report}a) demonstrate that
\ourMethodName~achieved the best performance across all ten metrics. Notably, compared with the second-best model (SimMIM~\cite{xie2022simmim}), \ourMethodName~outperformed by large margins of \(4.64\%\) (BLEU-4), \(9.43\%\) (accuracy) and \(18.70\%\) (F1-score).
Besides, the examples of generated reports (Fig. \ref{fig:fig_report}b) valid the ability of \ourMethodName~for ultrasound report generation, demonstrating the potential of \ourMethodName~in improving the efficiency, consistency, and accessibility of automatic clinical report generation.

\begin{figure}[!t]
    \centering
    \includegraphics[width=\linewidth]{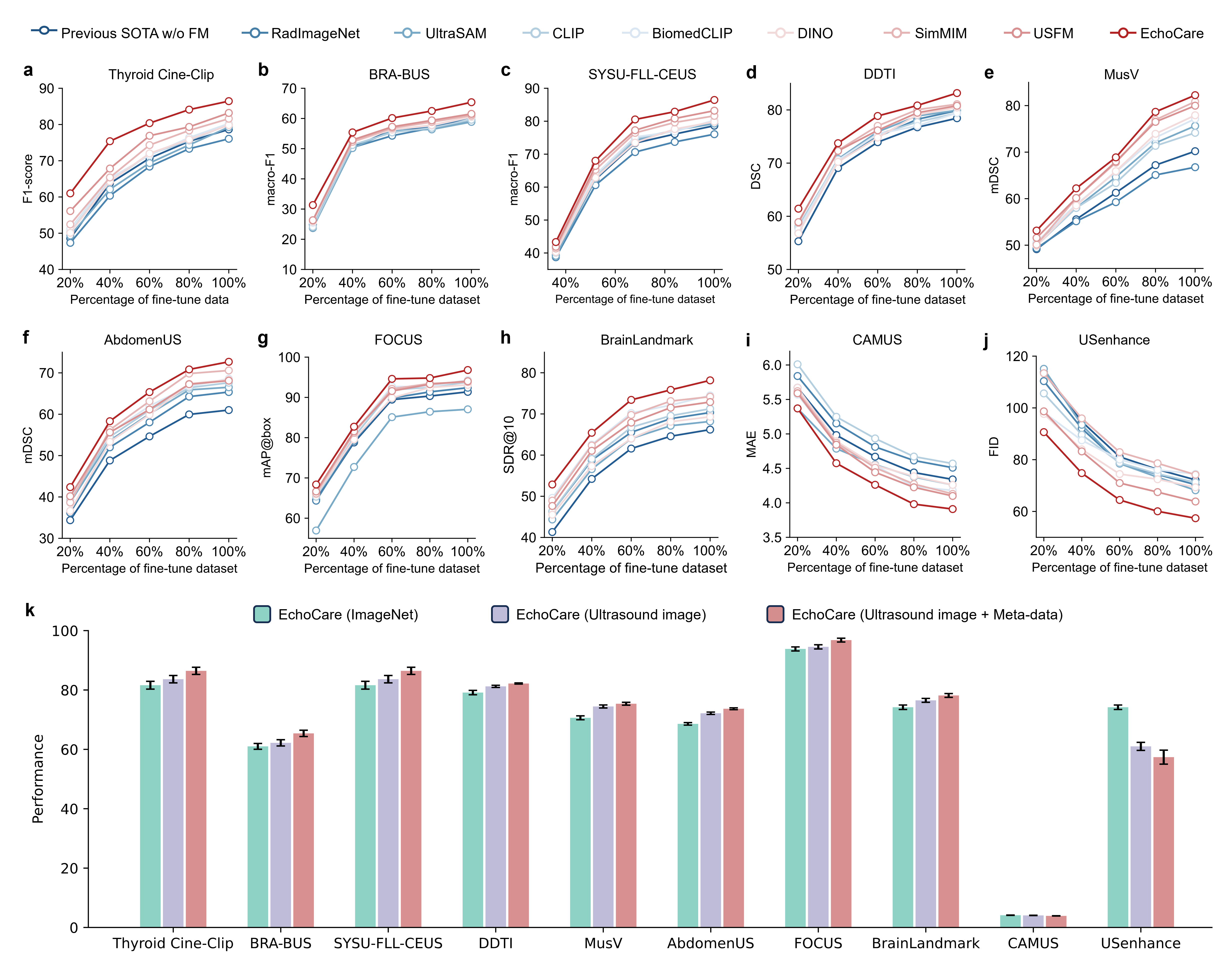}
    \caption{
     \textbf{Label efficiency and further analysis results.}
     \textbf{a-j.} Comparison between \ourMethodName~and other models (previous SOTA w/o FM and existing foundation models) in label efficiency generalization on ten clinical applications, showing performance at various training data percentages.
     \textbf{k.} Pretrained with large-scale ultrasound images, \ourMethodName~significantly improved performance (\%) over models based on natural image pretraining on ten clinical applications. Moreover, with our designed model dual-branch architecture, the performance can be further enhanced.
     }
    \label{fig:fig_ablation}
\end{figure}

\section{Discussion}
\label{sec:discussion}

Ultrasound imaging is a crucial tool in modern medicine. This work presents a novel, open-source foundation model named EchoCare to advance general-purpose clinical ultrasound applications. The model is pre-trained on our curated, publicly available dataset of 4.5 million ultrasound images, featuring a highly diverse and whole-body images sourced from over \countryNumber~countries and \continentNumber~continents. To evaluate EchoCare’s clinical utility, we conducted comprehensive validations across a wide range of downstream ultrasound tasks (lesion segmentation, disease diagnosis, organ detection, landmark prediction, quantitative regression, imaging enhancement and report generation). The results demonstrate the strong effectiveness and generalization capabilities of EchoCare: it consistently outperforms state-of-the-art foundation models such as UltraSAM~\cite{meyer2024ultrasam}, BiomedCLIP~\cite{zhang2025multimodal}, and USFM~\cite{jiao2024usfm} across all tasks.

The pre-training of our foundation model is powered by EchoCareData, the largest publicly available ultrasound image dataset to date, featuring over 4.5 million images. The vast scale and diverse of this dataset are crucial to our success. The core idea behind our data strategy is simple yet powerful: by aggregating numerous public datasets, we can significantly increase data size, expand protocol coverage, and diversify patient populations. This approach allows our model to learn from a broad spectrum of global sources. 
This extensive pre-training also grants EchoCare a remarkable degree of label efficiency for various downstream clinical tasks (Fig. \ref{fig:fig_ablation}a-j), thereby alleviating the substantial annotation workload for medical experts. For instance, in thyroid nodule segmentation, EchoCare can outperform other models using only 80\% of the labeled training data. Furthermore, \ourMethodName~showed consistently high adaptation efficiency, suggesting that \ourMethodName~required less time in adapting to downstream clinical applications, \eg,~ \ourMethodName~can potentially save about 20\% $\sim$ 40\% of the training time required to achieve convergence for the task of disease prediction. 

In addition to the substantial size and broad diversity of EchoCareData, the designed dual-branch architecture further contributes to the superior performance of \ourMethodName~across a wide range of downstream clinical tasks. Unlike previous medical foundation models that relied on standard MAE structures, our enhanced MAE architecture incorporates a unique anatomy-classifier branch. This branch is designed to learn global and hierarchical anatomical relationships, mirroring a clinician's diagnostic process. By integrating this high-level, representation-based learning with the local, pixel-level prediction of the MAE, our model's encoder gains a deeper understanding of ultrasound images. This dual-learning approach significantly boosts the model's ability to interpret images and perform well in a wide range of downstream clinical applications (Fig. \ref{fig:fig_ablation}k).

While \ourMethodName~has demonstrated promising potential of pretrained foundations for ultrasound analysis, several methodological frontiers remain. First, current pretraining exclusively uses image data, omitting clinically actionable text modalities (\eg, ultrasound diagnostic reports). Future iterations will integrate vision-language learning through curated datasets, enabling joint modeling of ultrasound images and associated clinical narratives to expand clinical applications. 
Second, \ourMethodName~currently treats dynamic modalities (\eg, videos) to static frames, thus failing to utilize the 
temporal cues essential for applications like cardiac motion analysis or vascular flow assessment. We will extend the architecture to incorporate spatio-temporal transformers, enabling end-to-end training on native video sequences and preserving temporal dynamics. Third, although results across 10 downstream clinical tasks demonstrate translational potential, rigorous validation is required before clinical adoption, such as real-world integration with clinical decision support systems.

In conclusion, we introduce EchoCare, a novel vision foundation model for ultrasound analysis, pretrained on our curated EchoCareData, which comprises over 4.5 million ultrasound images and is the largest ultrasound dataset to date. By integrating a novel architecture with a massive, diverse dataset, EchoCare establishes an efficient new paradigm for ultrasound image analysis, demonstrating robust adaptability to a broad spectrum of clinical ultrasound tasks and delivering significant performance gains over existing foundation models. Critically, we have made both the EchoCare model and EchoCareData publicly accessible to accelerate advancements in medical AI, improving clinical decision-making and patient care.

\clearpage
\section{Methods}\label{methods}
\subsection{Model design and pretraining}
For large-scale visual pretraining on \datasetName,
we proposed \ourMethodName,
a self-supervised framework for pre-training large vision transformer architectures based on the Masked Image Modeling (MIM) paradigm.
Specifically,
\ourMethodName~adopts a modular design,
comprising an image encoder, an image decoder, and a
meta-object classifier (Fig. \ref{fig:fig_overview}c), each module described in detail below.

The input to \ourMethodName~is a masked image, which is passed along to the image.
The image encoder processes the high-resolution image and outputs multi-scale downsampled embeddings.
We provide a flexible choice of backbone architectures with Swin Transformer base and large versions.
The image decoder outputs a reconstructed image that has the same size as the original image, with a grayscale value between 0 and 1 for each pixel.
The meta-object classifier includes input from the image and output object semantics.
The output object semantics includes three levels: part, organ and anatomical structure.
We follow SimMIM and SwinUNETR to build the image decoder head. The decoder is a transformer that gradually upsamples the image features back to high-resolution pixels. At the last layer, the attention dot product on the pixel embeddings delivers the reconstructed image.

\subsubsection{Unified masked pretraining}
The input image $x\in\mathrm{R}^{H\times W\times C}$ was split into $N$ image patches $\{x_i^p\}_{i=1}^N$ and then tokenized into 
$z = [z_1, ..., z_N] \in \mathrm{V}^{h\times w}$ as the output labels of MIM using an image patch embedding layer.
At the input layer,
50\% image patches were randomly masked, and then the model predicted the visual tokens $z_i$ of the masked patches.
Next, we replaced the masked patches with a learnable embedding $e_{[M]} \in \mathrm{R}^D$, making the input corrupted image patches $x^M = \{x_i^p: i \notin M\}_{i=1}^N \cup \{e_{[M]}: i \in M\}_{i=1}^N $ that are fed into the transformer encoder.
To optimize the model, we employ a reconstruction loss that aims to minimize the difference between the predicted pixel values of the masked image patches, $\hat{x}_i^p$, and the ground truth pixel values, $x_i^p$. Specifically, the reconstruction loss is defined as the Mean Absolute Error (MAE) between the predicted and original patches:

\begin{equation}
    \mathcal{L}_{\mathrm{MIM}} = \frac{1}{M}\sum_{i\in M} |\hat{x}_i^p - x_i^p|.
\end{equation}

\subsubsection{Hierarchical pretraining}
The second pre-trained output (\textit{i.e.}, the meta-objection classifier) is used to further train \ourMethodName~to represent images using hierarchical learning.
Therefore, we designed a hierarchical loss for image global representation learning.
Specifically, let’s assume there are $N_p$ body parts at the first level, which encompass $N_o$ organs at the second level. Based on these $N_o$ organs, there are $N_a$ anatomical structures at the third level.
Hence, the meta-object classifier has $N_p + N_o + N_a$ outputs.
For each category, if a class is labeled positive, all its ancestor nodes (\textit{i.e.}, superclasses) should be labeled positive.
And, if a class is labeled negative, all its child nodes (\textit{i.e.}, subclasses) should be labeled negative.
To ensure the satisfaction of the above hierarchy constraints, we estimate a hierarchy-coherent score vector
$P \in [0, 1]^{N_p + N_o + N_a}$.
For class $i$, the updated score vector $p = [p_i ] \in  [0, 1]$ in $P$ is given as:
\begin{equation}
    \begin{cases}
    p_{H} = \min(s_u)                        & \text{if } \hat{l} = 1, \\
    1 - p_{H} = \min(1-s_u) = 1 - \max(s_u)  & \text{if } \hat{l} = 0.
\end{cases}
\end{equation}
Thus, after getting the hierarchical probabilities, we could maximize the log-likelihood between the probabilities and ground truth classification labels:
\begin{equation}
    \mathcal{L}_{\mathrm{HIE}} = \sum -\hat{l}\log(p_H) - (1 - \hat{l})\log(1 - p_H).
\end{equation}

\subsection{Training settings}
Image augmentations included random vertical flip ($\textit{P}=0.5$), random horizontal flip ($\textit{P}=0.5$), and random crop ($\textit{P}=0.5$) to convert images to greyscale and weak colour jittering ($\textit{P}=0.2$) with specific adjustments to brightness, contrast, saturation and hue.
We pretrained \ourMethodName~for one million steps using the pretraining loss of $\mathcal{L}_{\mathrm{MIM}} + \mathcal{L}_{\mathrm{HIE}}$ for images.
The batch sizes were 256, and \ourMethodName~used an input image with $256\times256$ pixels and then patched as $2\times2$ pixels. 
We used the AdamW optimizer with $\beta_1=0.9$, $\beta_2=0.9$ and $\epsilon=0.9$ for optimization.
We used a cosine learning rate decay scheduler with a peak learning rate of $1.0\times10^{-4}$ and a linear warmup of 10,000 steps. The weight decay was set as 0.05, and the stochastic depth with a rate of 0.1 was used. 

\phantomsection
\subsection{Data curation for pretraining}\label{sec:data-curation}
Our data curation process commenced with a systematic search of open academic repositories, including Zenodo~\cite{Zenodo}, Mendeley~\cite{Mendeley}, Stanford AIMI, Figshare, and data/code platforms such as Kaggle~\cite{Kaggle}, GitHub~\cite{Github}, and medical challenge portals (\eg, Grand Challenge~\cite{Grand}). All data collection was concluded by 1 March 2025. Using "ultrasound" as a keyword, we retrieved approximately 13,000 potential datasets for initial screening. The raw dataset underwent a series of exclusion steps: 1) datasets were filtered to retain common file formats--including image files (\eg, PNG, JPG, BMP) and compressed archives (\eg, ZIP, RAR, TFRecord)--to confirm the presence of ultrasound images;
2) GPT-4o was utilized to extract direct download links from dataset descriptions in excluded text-only candidates;
3) preliminary deduplication was performed by comparing download URLs and computing image hash values for efficiency; and 4) manual curation was implemented to eliminate intra-organ redundancy through fine-grained filtering.
Moreover, to mitigate intrinsic anatomical sampling biases and ensure comprehensive coverage, we strategically prioritized underrepresented anatomical structures through targeted efforts: submitting formal access requests to specialized repositories (\eg, EchoNet-Dynamic) and directly contacting authors of ultrasound studies to procure supplementary datasets. Following our rigorous inclusion-exclusion pipeline, we compiled \datasetNumber~high-quality ultrasound datasets comprising over 4.5 million images, spanning nine major anatomical regions and 32 representative organs.

\subsection{Quality control and evaluation of the \datasetName~dataset}
An additional quality control and evaluation pipeline was implemented during construction of the \datasetName~dataset. To ensure data integrity, ultrasound images underwent a rigorous purification workflow: (1) Removal of extraneous patient metadata surrounding the image; (2) Discarding of completely empty images or those containing fewer than 1,000 valid (non-zero) pixels; (3) For ultrasound videos, systematic uniform sampling at 10-frame intervals to mitigate redundancy. Post-hoc evaluation of the filtering and deduplication processes was conducted as follows: after data filtering, a random sample of 100 excluded candidates was analyzed, confirming no valid ultrasound images or additional data links. Following deduplication, 100 potential duplicate datasets were manually assessed using a predefined similarity threshold ($\geq95\%$), verifying their redundancy. These procedures streamlined the dataset from 1,136 to 334 entries by eliminating redundancy. With the inclusion of specialized anatomically balanced datasets, we ultimately curated 138 high-quality ultrasound datasets.

\begin{table*}[!t]
\centering
\caption{Summary of the 10 clinical applications and the dataset distributions.}
\setlength{\tabcolsep}{1.5mm}      
\renewcommand{\arraystretch}{1.25}
\label{tab:application-dataset}
\resizebox{\textwidth}{!}{
\begin{tabular}{lllccccccc}
\toprule
Name & Challenge & Task type & Country & Anatomy part & Anatomy organ  & Train & Test \\
\midrule
DDTI~\cite{pedraza2015open}             & \makecell[l]{Thyroid node segmentation}      & Node segmentation & Colombia  &  Head \& Neck      & \makecell{Thyroid nodule}   &  522    &  123    \\
\midrule
MusV~\cite{geng2024force}             & \makecell[l]{Vessel segmentation}            & Vessel segmentation &  China      &  \makecell{Head \& Neck,\\Low limb}   & \makecell{Carotid and femoral vessels}    &  2,203    & 911    \\
\midrule
AbdomenUS        & \makecell[l]{Abdomen organ segmentation}     & Organ segmentation &  China      &  Abdomen         & \makecell{Liver, Kidney,\\Pancreas, Bladder, Spleen}   &  3,345    &  872    \\
\midrule
Thyroid Cine-Clip~\cite{Cine-Clip} & \makecell[l]{Thyroid node diagnose}        & Node classification &  USA   &  Head\&Neck      & \makecell{Thyroid nodule}   &  157    &  35    \\
\midrule
BRA-BUS~\cite{article}          & \makecell[l]{BI-RADS category\\assessment}    & BI-RADS classficaion & Brazil  &  Thorax          & \makecell{Breast}   &  1,500    &  375    \\
\midrule
SYSU-FLL-CEUS~\cite{article1}     & \makecell[l]{Liver lesion recognition}           & Lesion classification & China    &  Abdomen         & \makecell{Liver}   &  285    & 68   \\
\midrule
FOCUS~\cite{Wu2025}            & \makecell[l]{Thorax and cardiac\\organ detection} & Organ detection     & Spain    &  Fetus          & \makecell{Fetal thorax and cardiac}   &  250    &  50    \\
\midrule
BrainLandmark~\cite{cabezas2024benchmark}   & \makecell[l]{Brain landmark location}          & Landmark location  & Australia   &  Fetus     & \makecell{Fetal brain}   &  80    &  24    \\
\midrule
CAMUS~\cite{leclerc2019deep}           & \makecell[l]{Ejection fraction regression} & EF regression & France   &  Thorax     & \makecell{Heart}   &  450    &  50    \\
\midrule
USenhance~\cite{Guo2023}      & \makecell[l]{Low-quality image\\enhancement}    & Image enhancement &  China   & \makecell{Head \& Neck,\\Abdomen, Thorax}    & \makecell{Thyroid, Kidney, Liver,\\Breast, Carotid artery}   &  1,654  &  426    \\
\midrule
USreport~\cite{li2024ultrasound}      & \makecell[l]{Clinical report generation}      & Report generation &  China   &  \makecell{Head \& Neck,\\Abdomen, Thorax}     & \makecell{Thyroid, Liver, Breast}   &  1,118    &  279    \\
\bottomrule
\end{tabular}
}
\end{table*}

\subsection*{External validation tasks and benchmark datasets}\label{sec:validation-dataset}
To validate the generalizability of the pretrained foundation model \ourMethodName, we established 10 external validation tasks spanning representative clinical ultrasound scenarios including lesion segmentation, disease diagnosis, one-shot recognition, and quantitative regression. These tasks leveraged independent datasets covering anatomical regions such as thyroid, venous systems, abdominal organs, and cardiac structures. All external datasets were explicitly excluded from the \datasetName~pretraining corpus to prevent data leakage and ensure unbiased evaluation of pretraining effects. Below is a detailed breakdown of each clinical validation task and corresponding dataset, organized by task category to highlight translational relevance.

\myparagraph{Thyroid node segmentation on DDTI dataset (1 classes):}
The DDTI dataset~\cite{pedraza2015open} for thyroid node segmentation comprises 388 patients with B-mode ultrasound scans from the Instituto de Diagn\'ostico M\'edico S.A. and National University of Colombia, annotated for nodule lesion segmentation. Images were extracted from thyroid ultrasound video sequences acquired using TOSHIBA Nemio 30 and Nemio MX systems, equipped with 12 MHz convex and linear transducers. Accurate automated segmentation of thyroid nodules enables clinicians to assess morphological features--including size, shape, and margins--to discriminate between benign and malignant lesions, which is critical for early thyroid disease diagnosis. Sub-images were cropped from 42 composite sequences and integrated with single-frame images, yielding a total of 645 ultrasound images with an average resolution of $348\times280$ pixels and mean mask area of 153.25 pixels. For cross-validation, data were split at the patient level in an 8:2 ratio, resulting in 308:80 patient folds (522:123 images) for training and evaluation.

\myparagraph{Artery\&vein segmentation on Mus-V dataset (2 classes):}
The Mus-V dataset~\cite{geng2024force} for vascular segmentation comprises 3,114 ultrasound images from the Institute of Automation, Chinese Academy of Sciences, annotated for carotid and femoral vessel segmentation. Images were acquired from 11 healthy volunteers using an Angel Pionner H20 Ultrasound Scanner, capturing carotid and femoral vessels in the arm and neck regions. Accurate arterial-venous segmentation is critical for real-time low-risk vascular interventions--such as those for coronary and peripheral vascular diseases--enabling clinicians to precisely target vessels and minimize the risk of adjacent structure injury. The dataset includes separate annotations for arteries and veins to facilitate vascular analysis and identification, with images sampled from 105 videos (5-160 frames per video) at $400\times600$ pixel resolution. For evaluation, official train-test splits were used to achieve an 8:2 patient-level division, yielding 2,203:911 images for training and validation.

\myparagraph{Abdominal multi-organ segmentation on AbdomenUS dataset (5 classes):} Beyond single/two-class segmentation, \ourMethodName~was further validated on multi-organ segmentation to demonstrate its potential in reducing annotation burdens on experts. The AbdomenUS dataset for multi-organ segmentation encompasses 4,217 ultrasound images from BGI Genomics Co., Ltd., acquired from 64 volunteers using the MGIUS-R3 ultrasound system. Images were annotated for at least one of five abdominal organs: 1) liver, 2) pancreas, 3) kidney, 4) bladder, and 5) spleen. This multi-organ annotation framework allows clinicians to systematically evaluate anatomical morphology--including organ shape, positional relationships, and pathological signs--from diverse sonographic perspectives. For model training and validation, data were divided into an 8:2 ratio at the case level, yielding a training set of 51 cases (3,345 B-mode images) and a validation set of 13 cases (872 B-mode images).

\myparagraph{Thyroid nodule false positive mitigation on ultrasound cine-clip dataset (2 classes):}
The thyroid nodule false positive mitigation task leverages the Ultrasound Cine-clip dataset~\cite{Cine-Clip} from the Center for Artificial Intelligence in Medicine \& Imaging, comprising 192 histopathologically confirmed thyroid nodules (175 benign, 17 malignant) across 167 patients (mean age $56 \pm 16$ years, 137 female) who underwent cine ultrasound between April 2017 and May 2018. The dataset includes ultrasound cine-clip sequences, radiologist-annotated segmentation, patient demographics, lesion metrics (size/location), and definitive histopathological diagnoses.
Given the nonspecific nature of ultrasound findings, which often lead to unnecessary biopsies, AI-driven pre-biopsy triage of benign and malignant nodules holds significant clinical value for reducing false positive cancer classifications. All ultrasound acquisitions were performed using Logiq E9 (GE Healthcare) or Siemens S2000 systems, with images obtained by certified sonographers from supine patients with slightly hyperextended necks. The cine-clips feature $802\times1054$ pixel resolution. Following official dataset splits, the cohort was partitioned into training (157 cine-clips, 4/5) and validation (35 cine-clips, 1/5) subsets to ensure reproducible evaluation.

\myparagraph{BI-RADS category assessment on BRA-BUS dataset (4 classes):}
The Breast Imaging Reporting \& Data System (BI-RADS) category assessment leverages the BRA-BUS dataset~\cite{article}, which offers a standardized lexicon and reporting framework for breast ultrasound. BI-RADS facilitates consistent communication of imaging findings among radiologists and clinicians, with final assessments categorized by malignancy likelihood: categories 2 (benign), 3 (probably benign), 4 (suspicious), and 5 (highly suggestive of malignancy), as annotated by senior ultrasonographers.
The BRA-BUS dataset comprises 1,875 anonymized images from 1,064 female patients, acquired using four ultrasound systems (GE Logiq 5, GE Logiq 7, Toshiba Aplio 300, GE U-Systems) with linear-array transducers at the National Institute of Cancer (Rio de Janeiro, Brazil). For validation, an official 5-fold cross-validation strategy was employed, combining four folds into the training set (800 patients, 1,500 images) and using the remaining fold for validation (264 patients, 375 images).

\myparagraph{Focal liver lesion diagnosis on SYSU-FLL-CEUS dataset (3 classes):}
The focal liver lesion (FLL) diagnosis task leverages the SYSU-FLL-CEUS dataset~\cite{article1}, encompassing contrast-enhanced ultrasound data for three pathological types: 186 hepatocellular carcinoma (HCC), 109 hemangioma (HEM), and 58 focal nodular hyperplasia (FNH) cases. Acquired from the First Affiliated Hospital of Sun Yat-sen University using an Aplio SSA-770A ultrasound system (Toshiba Medical Systems), the dataset captures FLLs with heterogeneous patterns, varying in size, contrast intensity, morphological features, and anatomical location (resolution: $768\times576$ pixels). Early FLL characterization from ultrasound is critical for timely oncological intervention, as these lesions exhibit diverse imaging phenotypes. The dataset was case- and label-stratified into 8:2 training-evaluation folds to maintain class distribution: the training set includes 150 HCC, 88 HEM, and 47 FNH cases, while the evaluation set contains 36 HCC, 21 HEM, and 11 FNH cases.

\myparagraph{Fetal thorax and cardiac detection on FOCUS dataset (2 objects):}
The FOCUS dataset~\cite{Wu2025} is designed for fetal thorax and cardiac organ detection, comprising 300 four-chamber view fetal echocardiography ultrasound images from 217 subjects across Hospital Clinic and Hospital Sant Joan de Deu in Barcelona, Spain. This dataset captures the cardiothoracic diameter ratio--a critical biometric for assessing fetal congenital heart disease--via ellipse annotations of cardiac and thoracic regions in every image. All images ($230\times245$ pixels, uniform resolution) feature distinct annotations for fetal cardiac and thoracic structures, varying in size, aspect ratio, and rotational orientation. Following official patient-level splits to prevent data leakage, the dataset was partitioned into 250 training images and 50 evaluation images, maintaining clinical representativeness.

\myparagraph{Brain landmark detection on BrainBenchmark dataset (24 landmarks):} The brain landmark detection task leverages the BrainBenchmark dataset~\cite{cabezas2024benchmark}, comprising 104 2D fetal brain ultrasound images acquired at 20–20.6 weeks of gestation. Developed for monitoring neurodevelopmental trajectories, this benchmark captures structural changes from embryonic stages to postnatal development, with images obtained from 70 pregnant women (median age 31 years, range 18–42) via routine mid-trimester scans using a Voluson E10 ultrasound system with a high-frequency transabdominal probe (C2-9). Each image is annotated with 24 anatomical landmarks including 4 skull landmarks, 3 thalamic landmarks, 8 cerebellar perimeter landmarks, 4 cavum landmarks, 3 Sylvian fissure landmarks, and 2 midline edge landmarks. Images were collected from 70 subjects with variable scanning frequencies (8 women scanned three times, 18 women twice, and 44 women once), all without detected abnormalities. For validation, an 8:2 image-level split yielded 80 training and 24 evaluation images to ensure developmental stage representativeness.

\myparagraph{Ejection fraction prediction on CAMUS dataset (500 cases):} 
The CAMUS dataset~\cite{leclerc2019deep} for ejection fraction prediction comprises 500 2D ultrasound sequences, recognized as a standard benchmark for cardiac function assessment. This regression task involves inputting ultrasound frame sequences to predict left ventricular ejection fraction (LVEF), a critical biomarker for evaluating cardiac health and diagnosing heart disease, particularly when derived from four-chamber view acquisitions.
Ultrasound sequences were acquired using GE Vivid E95 scanners (GE Vingmed Ultrasound, Horten, Norway) with a GE M5S probe (GE Healthcare, US) at the University Hospital of St Etienne (France). Each sequence includes manual annotations of left ventricular volumes at end-diastole and end-systole, from which ejection fraction is calculated. Following official protocols, the dataset was partitioned into 450 training and 50 validation cases to ensure reproducible evaluation of LVEF prediction models.

\myparagraph{Image enhancement based on USenhance dataset (5 organs):}
The ultrasound image enhancement task~\cite{Guo2023} leverages the USenhance Challenge 2023 dataset, comprising 2,100 ultrasound images (1,050 unpaired low/high-quality image pairs) across five organs (thyroid, kidney, liver, breast, and carotid artery) from 109 patients. AI-driven enhancement of high-quality ultrasound images from low-fidelity inputs obviates the need for hardware upgrades, driving technological innovation in ultrasound devices and enabling more precise clinical applications. The dataset includes images acquired using diverse imaging systems: thyroid imaging employs the mSonics MU1 (low-end) and Toshiba Aplio 500 (high-end); carotid artery and abdominal imaging use SSUN (low-end) and Toshiba Aplio 500 (high-end); breast imaging utilizes the mSonics MU1 (low-end) and Aixplorer system from SuperSonic Imaging (high-end). All images were resized to a uniform $256\times256$ pixel resolution. Following an organ-stratified 8:2 split, the dataset was partitioned into 837 training image pairs (232 thyroid, 161 breast, 97 kidney, 119 liver, 228 carotid) and 213 validation image pairs (59 thyroid, 41 breast, 25 kidney, 30 liver, 58 carotid), ensuring clinical representativeness across anatomical structures.

\myparagraph{Ultrasound text report generation on USreport dataset (3 organs):}
The USreport dataset~\cite{li2024ultrasound} is designed for ultrasound text report generation, comprising three independent clinical corpora of ultrasound image-text pairs covering breast, thyroid, and liver examinations. Specifically, it includes 3,534 breast, 2,460 thyroid, and 1,397 liver cases, all sourced from the ultrasonic department database of the PLA General Hospital. AI-driven automated report generation from ultrasound images holds promise to streamline clinical diagnostic workflows. Each report is associated with two representative images selected by clinicians, forming image-text pairs for model training. Following official data splits, an 8:2 train-validation partition was applied: Liver: 1,118 training, 279 validation cases;
Breast: 2,827 training, 707 validation cases;
Thyroid: 1,968 training, 492 validation cases.

\section{Statistical analysis}
For all experimental results, performance metrics are reported as mean $\pm$ standard deviation across 20 independent trials. For each evaluation task, two-sample t-tests were conducted between the best-performing model and all others, with statistical significance denoted by asterisks ($\textit{*}p < 0.05$, $\textit{**}p < 0.01$, $\textit{***}p < 0.001$). A two-sided $\text{P-value} < 0.05$ was considered statistically significant.  For the thyroid nodule false positive mitigation binary-classification task, accuracy, sensitivity, and specificity were determined using the optimal cut-off value derived from the ROC curve to maximize the Youden index (sensitivity $+$ specificity $-$ 1). All statistical analyses were performed using {Python} (version 3.10) and {MedCalc} (version 22.032).
Across all experimental settings, results are visualized via box plots (version 3.9.1) showing quartiles and whiskers at 1.5$\times$ interquartile range, based on 20 repeated runs to characterize model performance variability.

\section{Data collection and analysis}
All source image data were from publicly available datasets. We used Python (version 3.10) to curate and preprocess the image.

\section{Data and code availability}
The curated ultrasound images in EchoCareData are available at \url{https://echocare.cares-copilot.com/}. All public datasets used in this work are listed with detailed download links on the project homepage.
% \section{Code availability}
The pretrained foundation model EchoCare is publicly available at \url{https://github.com/CAIR-HKISI/EchoCare}, including source code, installation guidelines, model weights, example datasets, and downstream task evaluation scripts.

%\section{Competing Interests}
%The authors declare no competing interests.

\bibliographystyle{sn-mathphys} % We choose the "plain" reference style
% \bibliography{references} % Entries are in the refs.bib file
\clearpage

% \input{content/figures}

% \appendix
% \setcounter{figure}{0}
% \setcounter{table}{0}

% \input{content/supplementary}

\end{document}